\pdfoutput=1

\documentclass[11pt]{article}

\usepackage[]{ACL2023}

\usepackage{times}
\usepackage{latexsym}

\usepackage[T1]{fontenc}

\usepackage[utf8]{inputenc}

\usepackage{microtype}

\usepackage{inconsolata}

\usepackage{graphicx}

\title{Can we trust the evaluation on ChatGPT?}

\author{Rachith Aiyappa,$^a$ Jisun An,$^a$ Haewoon Kwak,$^a$ \and Yong-Yeol Ahn$^{a,b}$ \\
$^a$ Center for Complex Networks \& Systems,\\ Luddy School of Informatics, Computing \& Engineering \\
$^b$ Indiana University Network Science Institute\\
Indiana University, Bloomington, Indiana, USA, 47408\\
\texttt{\{racball,jisunan,hwkwak,yyahn\}@iu.edu}}

\begin{document}
\maketitle
\noindent
\textbf{Link to the paper published in ACL Trust NLP workshop 2023}: \url{https://aclanthology.org/2023.trustnlp-1.5/}
\\

\noindent
\textbf{BibTeX}
\color{red}
\small
\begin{verbatim}
@inproceedings{
aiyappa2024trustChatGPT,
title = "Can we trust the evaluation on ChatGPT?",
author = "Aiyappa, Rachith  and An, Jisun and Kwak, 
Haewoon  and Ahn, Yong-yeol",
booktitle = "Proceedings of the 3rd Workshop on 
Trustworthy Natural Language Processing
(TrustNLP 2023)",
month = jul,
year = "2023",
address = "Toronto, Canada",
publisher = "Association for Computational 
Linguistics",
url = "https://aclanthology.org/2023.trustnlp-1.5",
doi = "10.18653/v1/2023.trustnlp-1.5",
pages = "47--54",
}
\end{verbatim}

\color{black}
\begin{abstract}
ChatGPT, the first large language model with mass adoption, has demonstrated remarkable
performance in numerous natural language tasks. Despite its evident usefulness, evaluating
ChatGPT’s performance in diverse problem domains remains challenging due to the closed
nature of the model and its continuous updates via Reinforcement Learning from Human
Feedback (RLHF). We highlight the issue of data contamination in ChatGPT evaluations,
with a case study in stance detection. We discuss the challenge of preventing data contamination and ensuring fair model evaluation in the age of closed and continuously trained
models.
\end{abstract}

\section{Introduction}

ChatGPT~\cite{chatgpt} has become the most prominent and widely-adopted pre-trained large language model (LLM) thanks to its impressive capabilities to perform a plethora of natural language tasks and its public accessibility. 
Although significant concerns regarding LLMs, particularly their tendency to ``hallucinate'' (or ``making things up'') and generation of biased or harmful content in scale have been raised~\cite{bender2021dangers, alkaissi2023artificial}, ChatGPT is becoming a common tool not only for everyday tasks such as essay writing, translation, and summarization~\cite{taecharungroj2023can,patel2023chatgpt}, but also for more sophisticated tasks such as code generation, debugging~\cite{sobania2023analysis}, and mathematical problem-solving~\cite{frieder2023mathematical}. 
With more than 100 million users within two months after its launch~\cite{milmo2023chatgpt} and its abilities pass hard exams like bar exam~\cite{terwiesch} and medical licensing exam~\cite{kung2023performance}, ChatGPT has stirred public perception of AI and has been touted as the paradigm for the next-generation search engine and writing assistant, which is already being tested by Microsoft's Bing search and Office products~\cite{bing}. Beyond commercial interests, LLMs  are also being tested for assisting scientific research~\cite{stokel2023chatgpt,dowling2023chatgpt,van2023chatgpt,wu2023large}.  

Although OpenAI---the creators of ChatGPT---performed internal tests, they do not cover all problem domains. Although the excellent general performance of ChatGPT is evident, it is still important to quantitatively characterize its performance on specific tasks to better understand and contextualize the model. 
Note that, given that it is currently not possible for a user to fine-tune ChatGPT, one can only evaluate it with a few-shot/zero-shot setting---a highly desirable setting that requires close to no annotated data.
A recent study showed that although ChatGPT performs generally well in many tasks, it has different strengths and weaknesses for different tasks and does not tend to beat the SOTA models~\cite{kocon2023chatgpt}.

However, given that the ChatGPT is a \emph{closed} model without information about its training dataset and how it is currently being trained, there is a large loxodonta mammal in the room: \emph{how can we know whether ChatGPT has not been contaminated with the evaluation datasets?} 

Preventing data leakage (training-test contamination) is one of the most fundamental principles of machine learning because such leakage makes evaluation results unreliable.
It has been shown that LLMs can also be significantly affected by data leakage, both by the leakage of labels and even by the leakage of dataset without labels~\cite{min2022rethinking,brown2020language,gpt4}.
Given that the ChatGPT's training datasets are unknown and that ChatGPT is constantly updated, partly based on human inputs from more than 100 million users via Reinforcement Learning from Human Feedback (RLHF)~\cite{chatgpt}, it is impossible to ascertain the lack of data leakage, especially for the datasets that have been on the internet. 

As far as it has been known, ChatGPT is trained in a three-step process. First, an initial LLM (GPT 3/3.5) is fine-tuned in a supervised manner on a dataset curated by asking hired human annotators to write what they think is the desired output to prompts submitted to the OpenAI API.\footnote{Additional labeler-written prompts are included too.} Next, a set of prompts is sampled from a larger collection of prompts submitted to the OpenAI API. For each prompt in this set, the LLM produces multiple responses, which are then ranked by human annotators who are asked to indicate their preferred response. The second step then trains a reward model (RM) on this dataset of response-ranking pairs to mimic the human ranking. This step keeps the LLM frozen and solely trains the RM. Finally, the LLM is made to generate responses to a set of prompts, which were not included in the previous steps, but submitted to the OpenAI API nevertheless. The now-frozen RM is used as a reward function, and the LLM is further fine-tuned to maximize this reward using the Proximal Policy Optimization (PPO) algorithm~\cite{schulman2017proximal}. 

Thus, if OpenAI continuously updates its models, by using queries submitted by researchers who wanted to evaluate ChatGPT's performance on various Natural Language Processing (NLP) tasks, it is likely that ChatGPT is already contaminated with the test datasets of many NLP tasks, which can lead to \emph{performance overestimation} in NLP tasks. 
Such contamination has been documented in the training data of other language models~\cite{brown2020language,dodge2021documenting,carlini2020extracting}.\footnote{\url{https://archive.is/44RRa}} 

It is important to highlight a distinction between two kinds of contamination acknowledged in literature~\cite{dodge2021documenting}: (1) the case where both the task input and labels are leaked to the model via training versus (2) the case where just the input is exposed. The latter is surely a smaller concern. However, even without the correct labels, exposure to the text in the same domain has been documented to increase the performance of the model to the corresponding NLP task~\cite{min2022rethinking}. 
Although we do not have any documented evidence that the ground-truth output answers/labels of the NLP tasks were submitted to the platform and the ChatGPT model has been trained with such data, we cannot exclude such possibility either. 
The annotator-generated responses to queries submitted to OpenAI during the RLHF step could potentially match the input text with output labels of the right kind; it is not possible to ensure no one has exposed certain input-label pairs to the model, for instance, via a few-shot learning experiment.
Given that language models show competitive performance in classification tasks despite poorly labeled data~\cite{min2022rethinking,garg2022can}, we cannot discard the possibility that the RLHF pipeline might essentially be a weaker variant of type (1) contamination.

Here, we use a case study of a stance detection problem~\cite{kuccuk2020stance} to raise awareness on this issue of data leakage and ask a question about how we should approach the evaluation of closed models. Stance detection is a fundamental computational tool that is widely used across many disciplines, including political science and communication studies.
It refers to the task of extracting the standpoint (e.g., Favor, Against, or Neither) towards a target from a given text. The task becomes more challenging when the texts are from social media like Twitter because of the presence of abbreviations, hashtags, URLs, spelling errors, and the incoherent nature of tweets. 
Recent studies have claimed that ChatGPT outperforms most of the previous models proposed for this task~\cite{zhang2022would} on a few existing evaluation datasets, such as the SemEval 2016 Task6 dataset~\cite{mohammad2016semeval,mohammad2017stance} and P-stance~\cite{li2021p}, even in a zero-shot setting where the model was not fine-tuned on the task-specific training data.    

Can this result be due to the data leakage and contamination of the model? 
Could this study itself have contaminated the ChatGPT model?
Although it is not possible to definitely answer these questions, it is \emph{also impossible to rule out} the possibility of contamination without the model owners' in-depth analysis. 

Following its release on Nov $30^{th}\ 2022$, on Dec $15^{th}$ $2022$, Jan $9^{th}$, Jan $30^{th}$, Feb $9^{th}$, and Feb $13^{th}\ 2023$, ChatGPT has been updated multiple times.\footnote{ChatGPT release notes: \url{https://archive.is/wHtXl}} While most of these releases updated the model itself, it is our understanding that the February releases were about handling more users to the platform, optimizing for speed, and the offering of ChatGPT plus---a subscription plan which provides priority access to new features, and faster response times.\footnote{ChatGPT plus: \url{https://archive.is/U0UxY}} 
Given that there has been at least one study that evaluated ChatGPT's performance on stance detection tasks~\cite{zhang2022would}, and that newer versions of ChatGPT are more likely to be \emph{exposed} to past queries to the platform, an opportunity arises to test whether the performance of the newer versions of ChatGPT on stance detection has been substantially improved after the study by~\cite{zhang2022would}. 

As we will present below, we do see an overall improvement in the performance before and after the publication of the stance detection evaluation paper~\cite{zhang2022would}. 
Of course, there is an alternative explanation that the model simply got better. 
However, we would also like to note that OpenAI has been updating the model primarily to address the model's problematic behaviors by making it more restricted, which led to the observation, although largely anecdotal, that the model has become `less impressive.' 

\begin{figure*}[ht]
    \centering
    \includegraphics[width=1\textwidth, trim={0 2in 0 2in}]{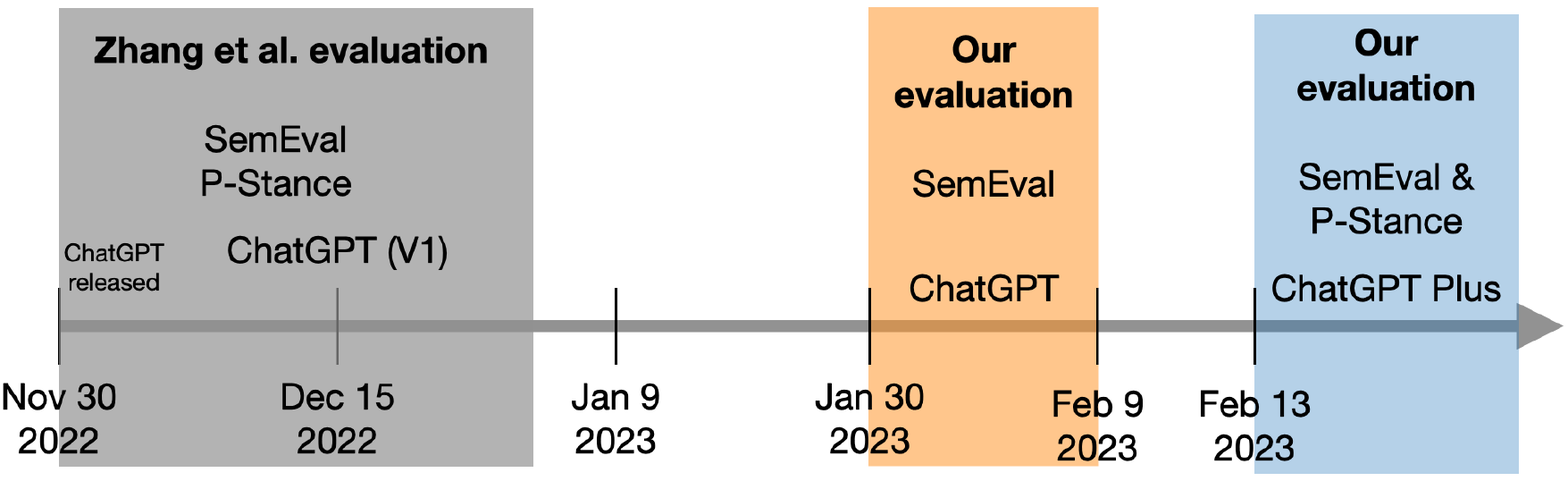}
    \caption{Updates of ChatGPT ever since its release on November 30, 2022. The versions of ChatGPT, each fine-tuned by RLHF process based on the queries to the OpenAI API platform, are indicated by the date ticks. The blocks contain the datasets, relevant to this study, on which ChatGPT's performance is evaluated on.}\label{fig:timeline}
\end{figure*}

\section{Methods}

Given that~\citealp{zhang2022would} was released on arXiv on December 30, 2022, and ChatGPT was launched on November 30, 2022, we assume~\citealp{zhang2022would} used either the November 30 or December 15 version of ChatGPT (henceforth called V1) to obtain their results (Fig.~\ref{fig:timeline}). Following their work, we used the test sets of SemEval 2016 Task 6~\cite{mohammad2016semeval,mohammad2017stance} and P-stance~\cite{li2021p} to perform our experiments. The
SemEval 2016 Task 6 dataset consists of relevant tweets in English with stance annotations towards six targets---`Hillary Clinton,' `Feminist Movement,' `Legalization of Abortion,' `Climate Change is a Real Concern,' `Atheism,' and `Donald Trump.' Similarly, the P-Stance dataset contains English tweets with stance annotations towards three targets---`Donald
Trump,' `Joe Biden,' and `Bernie Sanders.' 

We also used the same prompt. Specifically for SemEval 2016 Task 6, for instance, given the input: ``RT GunnJessica: Because i want young American women to be able to be proud of the 1st woman president \#SemST'', the input to ChatGPT is: ``\textit{what's the attitude of the sentence:} `RT GunnJessica: Because i want young American women to be able to be proud of the 1st woman president \#SemST' to the target `Hillary Clinton'. \textit{select from ``favor, against or neutral''}. 
Similarly, since the P-stance dataset does not have a neutral stance, the prompt is slightly modified to ``\textit{what's the attitude of the sentence:} `Air borne illnesses will only become more common with climate change. We need to immediately address this and fight for Medicare for All or this could be the new normal. \#BernieSanders' to the target `Bernie Sander'. \textit{select from ``favor, or against}''.~\footnote{This was confirmed with \citealp{zhang2022would} through email communication since the version of their paper at the time of writing this (\url{https://arxiv.org/pdf/2212.14548v2.pdf}) does not explicitly mention the prompt.}

Since ChatGPT did not provide an API to collect data at the time of the experiment, we first manually collected the responses of \textit{Jan 30th ChatGPT} for 860 tweets from the test data of SemEval 2016 Task 6, pertaining to the targets, `Hillary Clinton (HC),' `Feminist Movement (FM),' and `Legalization of Abortion (LA)' and extract the stance label from them. While the test set contains tweets pertaining to other targets (`Atheism,' `Donald Trump,' `Climate Change is a Real Concern'), we sampled the 860 tweets pertaining to the targets used in the previous work~\cite{zhang2022would}. 
After manual inspection of the preliminary results of the 860 tweets, we decided to collect and include the responses for the 2,157 tweets in the P-stance test dataset in our analysis, but the \textit{Jan 30th ChatGPT}  version was no longer available by then. Nevertheless, we use an open-source API~\footnote{\url{https://archive.is/6OGc3}} to automate the collection of responses from the \textit{Feb 13th ChatGPT plus} for both the P-stance and SemEval 2016 Task 6 datasets. Then we manually go through these (often verbose) responses to extract the stance labels from them when explicitly mentioned. 

In sum, we were only able to use the \textit{Feb 13th ChatGPT plus} version for the P-stance dataset and the \textit{Jan 30th ChatGPT} and \textit{Feb 13th ChatGPT plus} version for the SemEval 2016 Task 6 dataset because OpenAI (1) does not provide access to its older models after newer models are released, (2) imposes an upper bound on the number of requests which can be submitted to the platform in an hour, and, at the time of this experiment, (3) lacked a public API which in turn hindered the speed and efficiency of data collection. 

\section{Evaluation Metric and Results}

The macro-F and micro-F scores are shown for different versions of ChatGPT in a zero-shot setting on SemEval 2016 Task 6 and P-Stance datasets in Table~\ref{tab:sem} and Table~\ref{tab:pstance}, respectively. 
The macro-F score is calculated by averaging the F scores for the favor and against classes. 
The micro-F score is calculated by considering the total number of true positives, true negatives, false positives, and false negatives across the favor and against classes instead of averaging the F scores for each class. 

\begin{table*}
\centering
\begin{tabular}{llll}
\hline
\textbf{Model} & \textbf{HC} & \textbf{FM} & \textbf{LA}\\
\hline
V1 &  79.5/78.0 &  68.4/72.6 &  58.2/59.3 \\
Jan 30 ChatGPT  & \textbf{87.83/86.9} & \textbf{83.22/80.79} & \textbf{72.43/68.33} \\
Feb 13 ChatGPT plus &  82.9/81.87 & 75.94/71.96 & 65.56/61.74 \\
\hline
\end{tabular}
\caption{\label{tab:sem}Micro-F1/Macro-F1 scores of different versions of ChatGPT in a zero-shot setting on the SemEval 2016 Task 6 stance detection dataset.}
\end{table*}
\begin{figure*}
    \centering
    \includegraphics[width=1\textwidth, trim={0 3in 0 3in}]{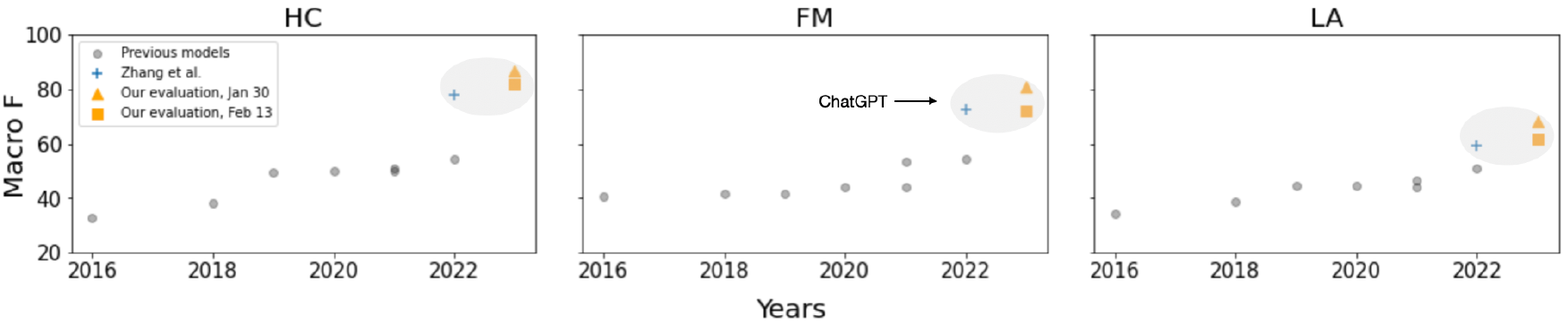}
    \caption{Evolution of zero-shot performance, measured using the macro-F score, on the SemEval 2016 Task 6A by various models. Scores of the previous models are taken from~\cite{zhang2022would}.}\label{fig:prev_zeroshot_semeval}
\end{figure*}
\begin{table*}
\centering
\begin{tabular}{llll}
\hline
\textbf{Model} & \textbf{Trump} & \textbf{Biden} & \textbf{Bernie}\\
\hline
V1 &  82.8/\textbf{83.2} &  82.3/82.0 &  79.4/79.4 \\
Feb 13 ChatGPT plus  & \textbf{83.76}/83.09 & 83.07/82.69 &  \textbf{79.7/79.6} \\
\hline
\end{tabular}
\caption{\label{tab:pstance}Micro-F1/Macro-F1 scores of different versions of ChatGPT in a zero-shot setting on the P-Stance stance detection dataset.}
\end{table*}
Overall, we see an improvement in performance, measured using the micro-F and macro-F scores, in recent versions of ChatGPT compared to V1. 
In particular, we see an average of 12.46 and 8.6 point improvement in the micro and macro-F scores, respectively, when comparing Jan 30 ChatGPT to V1 on the SemEval task. We see a smaller but non-negligible improvement---6.1 point on the micro-F and 1.89 point on the macro-F---when comparing Feb 13 ChatGPT plus to V1 on the same task. Fig.~\ref{fig:prev_zeroshot_semeval} also shows the temporal evolution of zero-shot performances of various models on selected targets of SemEval. The macro-F scores of the models are taken from the previous work~\cite{zhang2022would}. Although it is still difficult to conclude with only a few data points, we see a significant jump in the zero-shot capability of ChatGPT when compared to previous models. Given that ChatGPT is based on InstructGPT3 in which some NLP dataset contamination was already documented~\cite{brown2020language}, this raises further concerns if V1 too may have been contaminated.

A similar plot for the micro-F scores is not shown here due to our pending uncertainties of scores indicated in the previous work~\cite{zhang2022would} (see Appendix~\ref{app:lims}) and the general unavailability of micro-F scores by other models.
On the P-Stance dataset, we observe a 0.74-point improvement in the micro-F scores and a 0.26 point in the macro-F scores when comparing Feb 13 ChatGPT plus to V1.  

In sum, the improvement is greater for SemEval than for the P-Stance dataset. On the SemEval dataset, we also observe a performance drop by Feb 13 ChatGPT plus relative to Jan 30 ChatGPT. Even though the performance has dropped, it is still quite an improvement compared to V1.
\section{Discussion and Conclusion}~\label{sec:lims}
In this article, we discuss the reasons why we cannot trust the evaluation of ChatGPT models at its face value due to the possibility of data leakage. 
First, the closed nature of the model makes it impossible to verify whether any existing dataset was used or not.
Second, with a constant training loop, it is also impossible to verify that no researchers or users have leaked a particular dataset to the model, especially given the sheer scale of availability of the model (more than 100 million users\footnote{\url{https://archive.is/GiV3J}} at the time of writing). 
Any evaluation attempt using ChatGPT may \emph{expose} the very evaluation dataset to ChatGPT, potentially making all subsequent evaluations unreliable.
Note that even the mere exposure of the input may make evaluation unreliable~\cite{brown2020language,radford2019language}.
Therefore, unless the evaluation is completely novel, it is difficult to ensure the lack of data leakage to the model. 

Given that data leakage likely leads to a boost in estimated performance, we did a case study where there \textit{could} have been potential contamination, with documented evidence that researchers performed an evaluation of ChatGPT with an existing test dataset.  
In other words, the stance detection task that uses the SemEval 2016 Task 6 and P-stance datasets may no longer be a zero-shot problem for ChatGPT.
Although we cannot rule out the explanation that the ChatGPT is simply superior to previous models, it is also impossible to rule out the possibility of data leakage.  

This work sheds light on a bigger problem when it comes to using ChatGPT and similar large language models on NLP benchmarks. Given these models are trained on large chunks of the entire web, care must be taken to ensure that the pre-training and fine-tuning data of these models are not contaminated by the very benchmarks their performance is often tested on. Given the results showing that even benign contamination can lead to measurable differences, making claims about these models' zero-shot or few-shot inference capabilities require a more careful inspection of the training datasets of these models. For example, the BIG-bench dataset~\cite{srivastava2022beyond} attempts to address this issue by accompanying the benchmark data with a special string (``canary'' string). The purpose of this string is to allow researchers to better filter BIG-bench tasks out of the training data for large language models. This string also makes it possible to probe whether a language model was trained on BIG-bench tasks, by evaluating whether the model assigns anomalously high or low probabilities to the string.\footnote{BIG-bench canary string: \url{https://archive.is/CBgl2}}
Yet, checking for data contamination is becoming increasingly challenging because the most prominent language models, like ChatGPT and the recently released GPT-4,\footnote{GPT-4's technical report (\url{https://archive.is/9AucM}) says ``Given both the competitive landscape and the safety implications of large-scale models like GPT-4, this report contains no further details about the architecture (including model size), hardware, training compute, \textbf{dataset construction}, training method, or similar.''} are \emph{closed} and more models are following the practice. 

While our work is not without limitation (see `Limitations' section), we would like to underline that our primary goal of this article is to highlight the ample possibility of data leakage and the impossibility of verifying the \emph{lack of data leakage} with a closed model.
As long as the trend of closed models and continuous training loop continues, it will become more challenging to prevent data leakage (training-test data contamination) and ensure fair evaluation of models.
Therefore, in order to ensure the fair evaluability of the models, we argue that the model creators should (1) pay closer attention to the training datasets and document potential data contamination, (2) create mechanisms through which the training datasets and models can be scrutinized regarding data leakage, and (3) build systems that can prevent data contamination from user inputs.

\section{Data Availability}

The responses of ChatGPT, from which stance labels were manually extracted, are available upon request.

\section*{Limitations}~\label{sec:limitations}
Our analysis in this work is illustrative and exhibits many limitations. 
These limitations come from the fact that the ChatGPT system is new and being actively developed. 
The collection and extraction of stance labels from the responses of Jan 30 ChatGPT was done manually on the SemEval 2016 Task 6. However, due to the rate limitations, this was not done in one sitting since Jan 30 ChatGPT did not entertain more than a fixed (approx. 40) queries in an hour. There was a noticeable difference between the responses of ChatGPT at the beginning of the session (more verbose) when compared to when it was nearing its rate limit (less verbose; single-word responses). Additionally, in each sitting, a single chat session was used to feed multiple inputs, one at a time, to ChatGPT\footnote{sometimes factors like network errors which made ChatGPT unresponsive forced us to open a new chat session in the same sitting. But for a major chunk, a single session was used per sitting}, which may have accumulated context for subsequent inputs. In contrast, we used an open-source API for our experiments with the \textit{Feb 13 ChatGPT plus} version, which opened a new chat session per query. This may be one explanation for the drop in performance between Jan 30 and Feb 13 observed in Table~\ref{tab:sem} but recent work showed this to have an insignificant effect, although on a different dataset~\cite{kocon2023chatgpt}. An alternate explanation might be due to catastrophic forgetting---a documented phenomenon in large language models where the model tends to forget older information they were trained on in light of newer information~\cite{mccloskey1989catastrophic}. 
Yet another explanation could be that the Feb 13 ChatGPT plus is more \textit{diplomatic} than its predecessors given OpenAI's pursuit to make it less toxic and less biased. 
Due to the same reasons mentioned above, we could not try multiple queries for each input and could not estimate the uncertainty of the performance. 
The most critical limitation is, as we repeatedly stated above, that our result cannot prove nor disprove whether the data leakage happened or not as well as whether it has affected the evaluation of ChatGPT or not. 

\section*{Ethics Statement}
The findings of this work, though preliminary, and the problem of data contamination have major implications when it comes to using closed language models to conduct scientific research, measure progress in the field of natural language processing, and in commentaries about emergent properties/``intelligence'' of large language models.

Large language models are built on copious amounts of digital text which may contain sensitive and proprietary information.\footnote{\url{https://archive.is/DPcvj}} Methods and practices to ensure that this data is not included when creating language models are preliminary. Given the competitive landscape, and the trend of newer models being closed-source yet widely adopted, it is virtually impossible to verify the existence of such data in the training set. This calls for more efforts in designing experiments to quantify the presence and impact of such data, and methods to ensure that such data cannot be used/crawled. 

\section*{Acknowledgements}
We acknowledge the computing resources at Indiana University Bloomington and the ChatGPT platform by OpenAI which were used for the analysis in this paper. Yong-Yeol Ahn was supported by DARPA under contract HR001121C0168.

\bibliography{custom}
\bibliographystyle{acl_natbib}

\appendix

\section{Appendix}
\label{sec:appendix}

\subsection{Uncertainties in~\citealp{zhang2022would}}~\label{app:lims}
\
The results we obtain in Tables~\ref{tab:sem} and~\ref{tab:pstance} is compared against \citealp{zhang2022would}\footnote{\url{https://arxiv.org/pdf/2212.14548v2.pdf}} who used an older version of ChatGPT (called V1, in this paper). However, we believe that their work needs more clarification. At the time of writing this manuscript, we have requested further clarification from the authors.

The main source of uncertainty is the difference between the definitions of F1-m and F1-avg.~\citealp{zhang2022would} define F1-m to be the ``macro-F score" and F-avg as ``the average of F1 on Favor on Against" classes. It is our understanding that these two definitions are the same which would mean that for each target, the F1-m ad F1-avg should be the same. However, these scores are different from each other in~\citealp{zhang2022would}.
We also conjecture that there are a few misplaced scores in Tables 1, 2, and 3 in~\citealp{zhang2022would}. For instance, the scores of the PT-HCL and TPDG models in their Tables 1 and 2, should be the macro average F scores according to their original articles. However, these are placed under F1-avg and F1-m respectively in~\citealp{zhang2022would}.  
In our work, hoping to capture the \textit{worst case scenario}, we assume F1-m is the micro average and F1-avg is the macro average.  

Additionally, there is a mismatch between the input query to ChatGPT presented in the body of the previous work and that presented in the figures. We assumed that the format presented in the screenshot is what was used and selected it for this work with the neutral option being present (absent) for SemEval (P-Stance).

\end{document}